\pdfoutput=1

\documentclass[11pt]{article}

\usepackage[]{acl}

\usepackage{times}
\usepackage{latexsym}

\usepackage[T1]{fontenc}

\usepackage[utf8]{inputenc}
\usepackage{microtype}

\usepackage{amsmath,graphicx}
\usepackage{url}
\usepackage{multirow}
\usepackage{CJKutf8} 
\usepackage{textcomp }
\usepackage{comment} 
\usepackage{comment}
\usepackage[notextcomp]{stix}
\usepackage{xspace}
\usepackage{listings}
\usepackage{booktabs}
\usepackage{adjustbox}
\usepackage{enumitem}
\usepackage[toc,page]{appendix}
\usepackage{cleveref}
\usepackage{todonotes}
\usepackage{pifont}

\newcommand{\cmark}{\ding{51}}%
\newcommand{\xmark}{\ding{55}}%

\newcommand{\systemname}{\mbox{DOCmT5}\xspace}
\newcommand{\systemnameone}{\mbox{DOCmT5-5}\xspace}
\newcommand{\systemnametwo}{\mbox{DOCmT5-25}\xspace}
\newcommand{\targetname}{\mbox{DrMT}\xspace}
\title{\systemname: Document-Level Pretraining of Multilingual Language Models}
\newcommand\uw{$^{\diamondsuit}$}
\newcommand\ms{$^\spadesuit$}
\newcommand\aspace{\hspace{.75em}}

\author{
Chia-Hsuan Lee\uw\aspace
Aditya Siddhant\ms\aspace
 Viresh Ratnakar\ms\aspace 
  Melvin Johnson\ms\aspace \\
\uw University of Washington \aspace\ms Google Research \\
{\tt chiahlee@uw.edu}\\
{\tt \{adisid,vratnakar,melvinp\}@google.com}
}

\begin{document}
\maketitle
\begin{abstract}
In this paper, we introduce \textbf{\systemname}, a multilingual sequence-to-sequence language model pretrained with large scale parallel documents. While previous approaches have focused on leveraging sentence-level parallel data, we try to build a general-purpose pretrained model that can understand and generate long documents. We propose a simple and effective pretraining objective - \textbf{D}ocument \textbf{r}eordering \textbf{M}achine \textbf{T}ranslation (\textbf{\targetname}), in which the input documents that are shuffled and masked need to be translated. \targetname brings consistent improvements over strong baselines on a variety of document-level generation tasks, including over 12 BLEU points for seen-language-pair document-level MT, over 7 BLEU points for unseen-language-pair document-level MT and over 3 ROUGE-1 points for seen-language-pair cross-lingual summarization. We achieve state-of-the-art (SOTA) on WMT20 De-En and IWSLT15 Zh-En document translation tasks. We also conduct extensive analysis on various factors for document pretraining, including (1) the effects of pretraining data quality and (2) the effects of combining mono-lingual and cross-lingual pretraining. We plan to make our model checkpoints publicly available.
\end{abstract}


\section{Introduction}
\label{sec:introduction}

\begin{figure*}[t!]
    \centering
    \includegraphics[width=\linewidth]{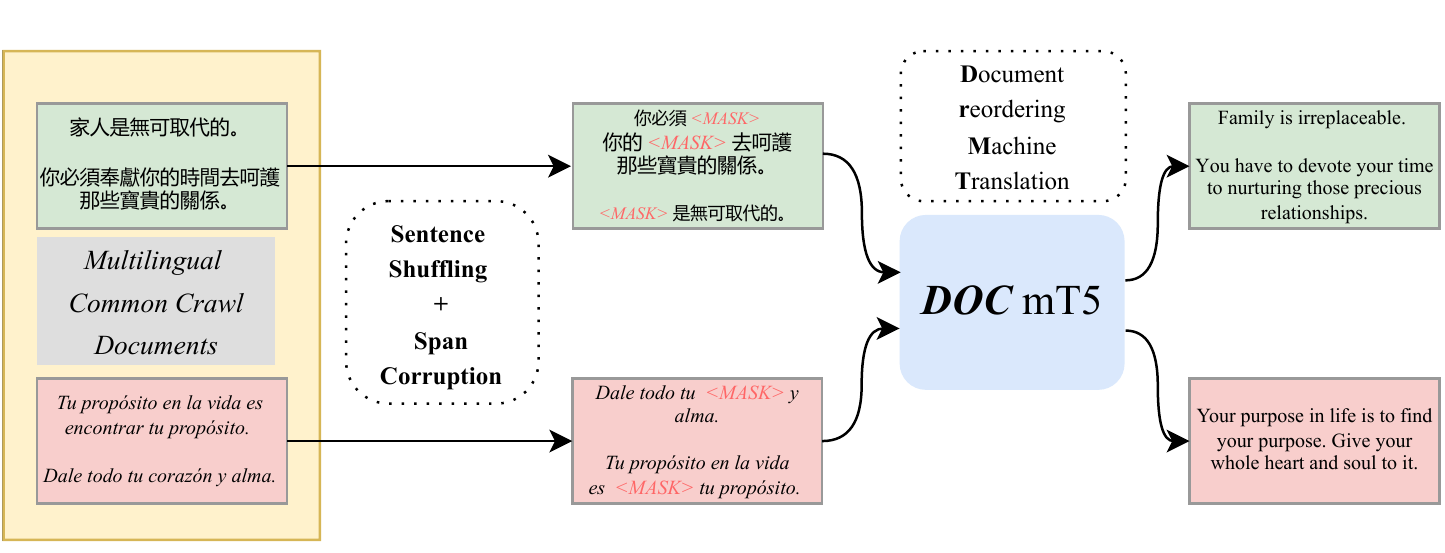}
    \caption{Overview of our proposed \textbf{D}ocument-\textbf{R}eordering \textbf{M}achine \textbf{T}ranslation (\textbf{\targetname}) pretraining. For each input document, the sentences are shuffled in random order and then randomly selected spans will be masked. The prediction target of \systemname is to generate the translation of the input document.}
    \label{fig:system}
\end{figure*}

Multilingual pretrained language models have been useful for a wide variety of NLP tasks. pretraining on large-scale  multilingual corpora facilitates transfer across languages and benefits low-resource languages.  

Previously, sentence-level or word-level cross-lingual objectives have been considered for pretraining large language models (LLM), but not much effort has been put in document-level objectives for pretraining. In this work, we propose a multilingual sequence-to-sequence language model pretrained with cross-lingual structure-aware document-level objectives. \systemname is built on top of mT5 \cite{xue2021mt5} and is further trained with parallel documents across multiple language pairs. To encourage the model to gain a deep understanding of the document structure and cross-lingual relationships, we consider a challenging translation scenario as a second-stage pretraining task: the input sentences are shuffled in a random order and random spans are masked. To effectively translate the input document, the model needs to reconstruct the document in the original order, making the model learn sentence relationships, and also recover the masked spans. This objective is effective on document-level generation tasks such as machine translation and cross-lingual summarization, outperforming previous best systems.

To enable cross-lingual pretraining at a large scale, we created a synthetic parallel document corpus. To avoid expensive human annotation, we use off-the-shelf neural machine translation (NMT) models to translate the documents in the mC4 corpus \cite{xue2021mt5} into English. In our experimental results, this corpus is more effective for pretraining than existing large-scale automatically aligned corpora (e.g., CCAligned \cite{el2020massive}).

We also conduct extensive ablation studies and provide insights on document-level pretraining. We show that simple document-level pretraining is more useful than sentence-level pretraining for generative tasks. We also show that data quality matters when performing multilingual document pretraining. Finally, we don't observe improvements from combining mono-lingual and cross-lingual objectives when evaluating on two document-level translation tasks. 

In summary, this paper makes the following contributions:
\begin{itemize}
    \item We build a state-of-the-art multilingual document-level sequence-to-sequence language model pretrained with a structure-aware cross-lingual objective. 
    \item Our proposed model achieves strong results on cross-lingual summarization and document-level machine translation for seen and unseen language paris, including SOTA on WMT20 De-En and IWSLT2015 Zh-En tasks.
    \item We also conduct extensive experiments to study what works and what doesn't work in document-level multilingual pretraining.
\end{itemize}

\section{Related Work}
\label{sec:related}

\subsection{Multilingual Pretraining}
Multilingual pretrained models provide a set of parameters that can be quickly finetuned for different downstream tasks~\cite{ruder-etal-2021-xtreme}. 
Some popular models are: mBERT \cite{devlin2019bert} and XLM-R \cite{conneau2020unsupervised} which pretrain with masked language modeling objective using only monolingual data, mT5 \cite{xue2021mt5} and mBART \cite{liu2020multilingual} which use a sequence-to-sequence language model and pretrain on large-scale mono-lingual corpora across many languages. Our proposed model uses mT5 as a backbone and further utilizes pseudo-parallel documents to learn better cross-lingual representations.

To capture cross-lingual information, translation language modeling \cite{conneau2019cross} and its variants (VECO \cite{luo2021veco}, ERNIE-M \cite{ouyang-etal-2021-ernie}) was proposed to leverage sentence-level parallel data. AMBER \cite{hu2021explicit} use two explicit alignment objectives that align representations at the word and sentence level. HICTL \cite{wei2020learning} pretrains on parallel sentences with word and sentence-level contrastive losses. mBART50 \cite{tang2021multilingual}, mT6 \cite{chi-etal-2021-mt6} and nmT5 \cite{kale-etal-2021-nmt5} focus on second-stage of pretraining using large-scale sentence-level translation data. Our model goes beyond the sentence and focuses on document-level understanding. 

While sentence-level pretraining has received a lot of attention, document-level pretraining has been under-studied. Unicoder \cite{huang-etal-2019-unicoder} replaces alternating sentences in a document with translations and pretrains with masked language modeling. MARGE \cite{lewis2020pre} adopts the retriever-generator paradigm and pretrains with an unsupervised translation objective on automatically retrieved documents. M2M100 \cite{fan2021beyond} pretrains sequence-to-sequence language models on automatically mined parallel sentences and documents. Our model considers a challenging supervised translation objective on parallel documents. 

\begin{table*}[ht]
    \small
    \centering

    \begin{tabular}{l@{\hskip5pt}c@{\hskip5pt}c@{\hskip5pt}c@{\hskip5pt}c@{\hskip5pt}c@{\hskip5pt}c}
    \toprule
    \textbf{Language} &  \textbf{Architecture}  &  \textbf{Parameters}  & \textbf{\# Languages} & \textbf{Monolingual Data} & \textbf{Cross-Lingual Data} & \textbf{Parallel Docs} \\
        \midrule
        mBERT & Encoder-only & 180M & 104 & Wikipedia  &  \xmark  &  \xmark  \\
        RemBERT & Encoder-only & 980M & 110 & Wikipedia and  Common Crawl  &  \xmark  &  \xmark  \\
        XLM  &  Encoder-only &  570M & 100 &  Wikipedia  & Misc. &  \xmark  \\
        XLM-R &  Encoder-only & 270M - 550M & 100 &  Common Crawl (CCNet) &  \xmark  &  \xmark  \\
        mBART & Encoder-decoder & 680M & 25 & Common Crawl (CC25)  &  \xmark  &  \xmark  \\
        mBART50 & Encoder-decoder & 680M & 50 & Common Crawl (CC25) & ML50 &  \textcolor{red}{\cmark} \\
        MARGE & Encoder-decoder & 960M  & 26 & Wikipedia or CC-News  &  \xmark  &  \xmark  \\
        mT5 & Encoder-decoder & 300M - 13B & 101 &  Common Crawl (mC4) &  \xmark  &  \xmark \\
        nmT5 & Encoder-decoder & 800M - 3B & 101 & Common Crawl (mC4) &  OPUS-100 & \xmark \\ 
        \midrule
        \textbf{\systemname (ours)} & Encoder-decoder & 580M - 800M  & 25 & Common Crawl (mC4)  & MTmC4 & \textcolor{red}{\cmark} \\
        \bottomrule
    \end{tabular}
        \caption{Comparisons of \systemname to previous multilingual language models.} 
    \label{tab:LLM}
\end{table*}

\begin{table}[ht]
    \small
    \centering

    \begin{tabular}{cc|cc}
    \toprule
     \textbf{Language}  & \textbf{Size/GB} & \textbf{Language}   & \textbf{Size/GB}     \\
        \midrule
        De$\star$  & 44  & Ar & 58 \\
        Es$\star$ &   52  &  Az & 42 \\
        Tr$\star$ & 45 &  Bn & 66 \\
        Ru$\star$ & 58 & Bn & 66 \\
        Vi$\star$ & 50 & Fa & 54 \\
        Fi & 47 &  Ko & 87 \\
        Fr & 43 & Lt & 48 \\
        Hi & 20 & Mr & 125 \\
        It & 40 & Nl & 38 \\
        Ja & 120 & Pl &  45 \\
        Pt & 40 &  Th & 63 \\
        Ro & 53 & Uk  & 66 \\
        Zh & 41 \\
        \bottomrule
    \end{tabular}
        \caption{Statistics of the MTmC4 corpus. $\star$ indicates that the language is used in \systemnameone.}  
    \label{tab:stats}
\end{table}
\vspace{-0.2cm}

\subsection{Multilingual Parallel Data Sources}
OPUS-100 \cite{aharoni2019massively,zhang2020improving} is collected from a variety of domains and is human labeled but it is at the sentence level. ML50 \cite{tang2021multilingual} is collected from different machine translation challenges and other publicly available corpora such as OPUS, but most of the data is at the sentence level. CCMatrix \cite{schwenk-etal-2021-ccmatrix} and Wikimatrix \cite{schwenk2021wikimatrix} use multilingual sentence embedding to automatically mine parallel sentences. Perhaps the most closest to our proposed corpus is CCAligned \cite{el2020massive}, which is also automatically mined but its quality is in question \cite{kreutzer2021quality}. Our MTmC4 corpus does not require human annotation and instead was produced by NMT models. 


\subsection{Document-level Machine Translation}
There are different ways to incorporate document context into translation model. Just to name a few, previous works have explored concatenation-based methods \cite{tiedemann2017neural,junczys2019microsoft,sun2020capturing,lopes-etal-2020-document}, multi-source context encoder \cite{zhang2018improving,jean2017does}, and  hierarchical networks \cite{zheng2020towards,zhang2020long,chen2020modeling}. This line of research focuses on architectural modifications of neural translation models. We focus on how to design a generalized pretraining objective and furthermore, our model can be finetuned for various downstream tasks (e.g. summarization) without task-specific changes.

\section{Multilingual Pretraining}
\label{sec:system}

\subsection{Datasets}
\subsubsection{mC4}
For pretraining, we use mC4 \cite{xue2021mt5}, a large scale corpus extracted from Common Crawl that covers over 100 languages.

\subsubsection{MTmC4: Creating Parallel Documents with mC4}
To create large-scale parallel documents, we take mC4 as a starting point and use in-house NMT models to translate documents from 25 languages into English. Each sentence in each document is translated independently. For each language, we sample 1 million documents, if there are more than that to start with, in mC4. Detailed data statistics for all the languages can be found in Table \ref{tab:stats}. 

\subsection{Document Reordering Machine Translation (\targetname)}
\label{sec:DrMT}
We start by introducing two related pretraining objectives: 
\begin{itemize}
    \item \textit{NMT Pretraining}: 
    \citet{tang2021multilingual} and \citet{kale-etal-2021-nmt5} proposed to perform a second-stage of pretraining using sentence-level MT data. The objective here is to perform sentence-level translation without any other changes to the input.
    
    \item \textit{Monolingual Document Reordering (Dr) Pretraining}:
    This objective, proposed by mBART \cite{liu2020multilingual}, changes the order of the sentences in each document. This is then followed by the original span corruption objective in T5. The decoder is required to generate the original document in order.
\end{itemize}

We combine these two objectives and propose \textbf{\targetname}. In \targetname, we introduce two types of noise on the input: \textbf{(i)} sentences in the document are randomly shuffled and \textbf{(ii)} randomly sampled spans are masked. In order to correctly translate the content, the model needs to decipher the corrupted document in order first. This enforces the models to gain deep understanding of the document structure. More formally, suppose we have N language pairs and each language has a set of parallel documents, the whole collection of document pairs are $D = \{ D_{1}, D_{2}, ..., D_{N}\} $. And a pair of $(x, y)$ is an instance in one of the language documents $D_{i}$. The overall learning objective is maximizing the likelihood of $y$ given a corrupted $C(x)$, that is \begin{equation} \sum_{D_{i}\in D} 
\sum_{(x, y) \in D_{i}} \log P(y|C(x)).
\end{equation}

\subsection{\systemname}
We use mT5 as the backbone model. mT5 is a sequence-to-sequence language model pretrained with the span corruption objective in which random spans in the input are masked and the decoder is required to reconstruct the masked spans (see \citet{raffel2020exploring} and \citet{xue2021mt5} for further details). Our system, \systemname, incorporates a second-stage pretraining with a structure-aware cross-lingual objective(\ref{sec:DrMT}) on pseudo parallel documents. Detailed comparisons with previous multilingual language models can be found in Table \ref{tab:LLM}.
We provide two variants of \systemname with both Base and Large model settings: 
\begin{itemize}
    \item \textbf{\systemnameone} This model is pretrained with 5 languages: \{De, Ru, Tr, Vi and Es\}. For all of the pretraining objective baselines in this paper, we pretrain with this set of languages, unless specified otherwise.
    \item \textbf{\systemnametwo} This model is pretrained with 25 languages. We show the full list of languages and their sizes in Table \ref{tab:stats}.
\end{itemize}

\subsection{Implementation Details}
We use mT5-Base\footnote{\url{https://console.cloud.google.com/storage/browser/t5-data/pretrained_models/mt5/base/}} and mT5-Large\footnote{\url{https://console.cloud.google.com/storage/browser/t5-data/pretrained_models/mt5/large/}} checkpoints at 1M steps as our pretrained models. We perform a second-stage of pretraining for an additional 0.5M steps using batches of 256 examples each of max length 1024. The learning rate is determined by a inverse square root scheduler as defined in T5, with the learning rate set to $ 1  / \sqrt{n}$ where n is the number of training step. We use the same span corruption objective as T5,
with 15\% of random tokens masked and an average noise span length of 3. For finetuning, we use a constant learning rate of 0.001 and dropout rate of 0.1 for all tasks until convergence. We adopt greedy decoding during inference.

\section{Experiments}
\label{sec:experiment}

\subsection{Baselines}
\begin{itemize}
    \item \textbf{Second-Stage Pretraining on 5 Languages} \\
    Language models pretrained with huge numbers of languages suffer from curse of multilinguality. In order to make a fair comparison, we create a strong mT5 model by continuing to pretrain on the same 5 languages of mC4 as in \systemnameone with the same number of steps using the original span corruption objective in mT5. Models pretrained with this objective is denoted as \textbf{cont-5langs}.
    
    \item \textbf{Monolingual Document Reordering (Dr)} \\
    We briefly mention this objective in Section\ref{sec:DrMT}. We use the mC4 corpus for this pretraining objective. Models pretrained with this objective is denoted as \textbf{Dr} (\textbf{D}ocument \textbf{R}eordering).
    
    \item \textbf{Document TLM (DocTLM)} \\
    In \citet{conneau2019cross}, the authors propose the translation language modeling(TLM) objective, which concatenates parallel sentences and applies masked language modeling to learn cross-lingual knowledge. Here we extend it to the document level by concatenating parallel documents. Instead of masking single tokens, we follow the span corruption objective in T5 and mask consecutive spans. The models are pretrained with this objective on MTmC4. 
    \item \textbf{Document NMT (DocNMT)} \\
    We consider a standard document-level machine translation for pretraining. The source document is the input and the target translation is the output. We use MTmC4 for this pretraining objective.

\end{itemize}

\begin{table*}[ht]
    \small
    \centering

    \begin{tabular}{l@{\hskip5pt}c@{\hskip5pt}c@{\hskip5pt}c@{\hskip5pt}c|@{\hskip5pt}c}
    \toprule
    \textbf{Pretrained Model} &  \textbf{Es-En}  &  \textbf{Ru-En}  &  \textbf{Tr-En}  &  \textbf{Vi-En}   &  \textbf{Average} \\
        \midrule
        \multicolumn{5}{c}{\textit{Previous Systems}} \\
        \midrule
        mBART & \textbf{38.30 / 15.40 / 32.40} &	33.10 / 11.90 / 27.80  & 34.40 / 13.00 / 28.10 & 32.00 / 11.10 / 26.40	 & 34.45 /	12.85 /	28.67  \\
        \midrule
        \multicolumn{5}{c}{\textit{Mono-Lingual}} \\
        \midrule
        mT5 & 29.97 / 10.65 / 25.70  & 27.91 / 8.90 / 22.60 & 29.98 / 11.96 / 24.56 &  24.38 / 7.39 / 19.59 & 28.06 /  9.72  / 23.11\\
        \quad \textit{w.} cont-5langs & 34.50 / 12.83 / 28.37 & 30.20 / 10.30 / 24.77 & 32.12 / 13.71 / 26.40 & 28.95 /  9.74 / 23.76 & 31.44 / 11.64 / 25.82 \\
        \quad \textit{w.} Dr  & 36.22 / 14.18 / 30.31 & 32.29 / 11.64 / 26.63 & 34.25 / 14.93 / 28.50 & 30.07 / 10.46 / 25.00  & 33.20 / 12.80 / 27.61 \\
        \midrule
        \multicolumn{5}{c}{\textit{Cross-Lingual}} \\
        \midrule
        \quad \textit{w.} DocNMT & 33.45 / 12.56 / 29.04 & 30.93 / 11.01 / 25.82  & 33.32 / 14.10 / 27.54 & 27.60 / 9.26 / 22.52 & 31.40 / 11.59  / 26.12\\
        \quad \textit{w.} DocTLM & 35.40/ 13.76 / 29.71 & 30.26 / 10.33 / 24.78 & 34.85 / 15.35 / 28.88 & 30.35 / 10.86 / 25.03  & 32.71 / 12.57 / 27.10 \\
        \midrule
       \systemnameone & 36.60 / 14.55 / 30.64 & 32.90 / 12.09 / 27.41 & 37.02 / 16.64 / 30.97 & 32.13 / 11.81 / 26.72 & 34.66 / 13.77 / 28.93\\
        \systemnameone-Large & 36.34 / 14.69 / 31.14 & 33.15 / 12.32 / 27.80 & 37.11 / 16.40 / 30.63 & \textbf{33.29 / 12.35 / 27.50} & 34.97 / 13.94 / 29.26\\
       \systemnametwo & 36.42 / 14.47 / 30.51 & 30.99 / 10.94 / 25.78 & 35.99 / 16.13 / 29.67 & 31.71 / 11.53 / 26.40 & 33.77 / 13.26 / 28.09\\
        \systemnametwo-Large & 36.79 / 15.04 / 31.48 &\textbf{33.56 / 12.77 / 28.46} & \textbf{37.66 / 16.68 / 31.37} & 32.43 / 11.87 / 27.04 & \textbf{35.11 / 14.09 / 29.58} \\
        \bottomrule
    \end{tabular}
        \caption{Results of four seen langauges paris \{Es, Tr, Ru, Vi\} on Wikilingua. Each cell demonstrates three metrics: ROUGE-1, ROUGE-2 and ROUGE-L in order. The mBART results are taken from the GEM\cite{gehrmann-etal-2021-gem} paper for a strong baseline model.} 
    \label{tab:DR-summary}
\end{table*}

\begin{table*}[ht]
    \small
    \centering
    \begin{tabular}{l@{\hskip5pt}c@{\hskip5pt}c@{\hskip5pt}c|@{\hskip5pt}c}
    \toprule
    \textbf{Pretrained Model} &  \textbf{Fr-En}  &  \textbf{Id-En}  &  \textbf{Hi-En}     &  \textbf{Average} \\
        \midrule
        \multicolumn{5}{c}{\textit{Mono-Lingual}} \\
        \midrule
        mT5 & 29.66 /  9.96 / 24.37 & 29.08 / 9.87 / 23.83 & 26.18 / 8.51 / 20.91 &  28.30 / 9.44 / 23.03 \\
        \quad \textit{w.} cont-5langs & 32.78 / 11.79 / 27.29 & 32.21 / 11.65 / 26.36 & 28.93 / 10.06 / 23.37 & 31.30 / 11.16 / 25.67  \\
        \quad \textit{w.} Dr  & 34.47 / 12.67 / 28.58 & 34.05 / 12.87 / 27.96 & 31.13 / 11.18 / 25.16 &  33.21 / 12.24 / 27.23  \\
        \midrule
        \multicolumn{5}{c}{\textit{Cross-Lingual}} \\
        \midrule
        \quad \textit{w.} DocNMT & 33.22 / 12.33 / 27.97 & 31.97 / 11.80 / 27.11  & 29.33 / 10.12 / 23.86 & 31.50 / 11.41 / 26.31  \\
        \quad \textit{w.} DocTLM & 32.79 / 11.75 / 27.12  & 33.35 / 12.24 / 27.37  & 30.48 / 11.24 / 24.92 & 32.20 / 11.74 / 26.47  \\
        \midrule
        \systemnameone & 34.02 / 12.57 / 28.21 & 34.31 / 13.09 / 28.56 & 32.24 / 11.84 / 26.06 & 33.52 / 12.50 / 27.61 \\
        \systemnameone-Large & \textbf{36.28 / 14.27 / 30.78} & 34.52 / 13.45 / 29.22 & 33.15 / 12.68 / 27.35 & 34.65 / 13.46 / 29.11 \\
       \systemnametwo &  34.56 / 13.10 / 29.03  & 34.16 / 13.04 / 28.23 & 32.33 / 11.99 / 26.25 & 33.68 / 12.71 / 27.83 \\       
       \systemnametwo-Large &  35.66 / 13.99 / 30.26  & \textbf{35.15 / 13.70 / 29.47} & \textbf{34.16 / 13.26 / 27.93} & \textbf{34.99 / 13.65 / 29.22} \\
        \bottomrule
    \end{tabular}
        \caption{Results of three unseen langauges paris \{Fr, Id, Hi\} on Wikilingua.} 
    \label{tab:DR-summary-unseen}
\end{table*}

\subsection{Cross-Lingual Summarization} 
We evaluate \textit{\systemname} on cross-lingual summarization as it is challenging for the model to summarize a long document and translate the salient information at the same time. We use Wikilingua, a cross-lingual summarization dataset, in which a document from an arbitrary language must be summarized in English. We adopt the GEM \cite{gehrmann-etal-2021-gem} version where the data is re-split to avoid train-test overlap between languages. We use a special prefix for cross-lingual summarization: \textit{"Summarize X to Y"}, where X and Y are the source and target language names respectively.

\subsubsection{Results on Seen Language Pairs}
We show the finetuning results of language pairs that are in the second stage of pretraining in Table ~\ref{tab:DR-summary}. We use the same four languages that are in Wikilingua's original release \{Es, Ru, Tr, Vi\}. The \textit{Dr} objective brings substantial improvements over \textit{cont-5langs} in all four languages, justifying the importance of structure-aware objectives. As for cross-lingual objectives, \textit{DocTLM} is better than \textit{DocNMT} in almost all languages except for Russian. \systemnameone substantially outperforms \textit{DocNMT} and \textit{DocTLM}, showing that our proposed pretraining objective leads to improved cross-lingual learning. The results of \textit{\systemnametwo} are inferior to \textit{\systemnameone} and this is possibly due to capacity dilution \cite{arivazhagan2019massively}. As we increase the capacity, we see that \textit{\systemnametwo-Large} outperforms \textit{\systemnameone-Large}. \textit{\systemnametwo-Large} is the best overall model outperforming the strong prior system: mBART.

\subsubsection{Results on Unseen Language Pairs}
We show the finetuning results of language pairs that are not in the second-stage of pretraining stage in Table \ref{tab:DR-summary-unseen}. We use three languages \{Fr, Id, Hi\}\footnote{We choose French to study the transfer ability of the cross-lingual models on high-resource and same-script (latin) languages. Indonesian is for studying high-resource and different-script language. Hindi is for studying low-resource and different-script language.}. Once again, we see that the \textit{Dr} objective brings substantial improvements over \textit{cont-5langs}. Surprisingly, without directly pretraining on the same language pairs, \textit{\systemnameone} leads to substantial improvements over strong baselines. This shows that our pretraining objectives are able to generalize to other languages. \textit{\systemnametwo} pretrains on French and Hindi but not Indonesian and hence we observe improvements of average results over \textit{\systemnameone}. The improvements of \textit{\systemname} are not so substantial and sometimes even hurt performance in high-resource languages: French and Indonesian, which have 44556 and 33237 training examples respectively and there are only 6942 examples in Hindi. \textit{\systemnametwo-Large} obtains the best results in almost all 3 languages except for French.

\begin{table}[ht]
    \small
    \centering

    \begin{tabular}{l@{\hskip7pt}c}
    \toprule
    \textbf{Pretrained Model} &  \textbf{d-BLEU} \\
        \midrule
        \multicolumn{2}{c}{\textit{Previous Systems}} \\
        \midrule
        NTT \cite{kiyono2020tohoku} & 43.80 \\
        PROMT \cite{molchanov2020promt} & 39.60\\
        OPPO \cite{shi-etal-2020-oppos} & 42.20 \\
        \midrule
        \multicolumn{2}{c}{\textit{Mono-Lingual}} \\
        \midrule
        mT5 & 29.08 \\
        \quad \textit{w.} cont-5langs & 32.24 \\
        \quad \textit{w.} Dr  & 36.71\\
        \midrule
        \multicolumn{2}{c}{\textit{Cross-Lingual}} \\
        \midrule
        \quad \textit{w.} DocNMT & 41.23\\
        \quad \textit{w.} DocTLM & 37.74 \\
        \systemnameone & 42.19 \\
        \systemnameone-Large & \textbf{44.73} \\ 
        \systemnametwo & 40.99 \\
        \systemnametwo-Large & 43.49\\
        \bottomrule
    \end{tabular}
        \caption{Finetuning results on WMT20 De-En.} 
    \label{tab:DR-WMT20}
\end{table}

\begin{table}[ht]
    \small
    \centering

    \begin{tabular}{l@{\hskip7pt}c}
    \toprule
    \textbf{Pretrained Model} &  \textbf{d-BLEU} \\
        \midrule
        \multicolumn{2}{c}{\textit{Previous Systems}} \\
        \midrule
        HAN & 24.00 \\
        mBART & 29.60 \\
        MARGE & 28.40 \\
        \midrule
        \multicolumn{2}{c}{\textit{Mono-Lingual}} \\
        \midrule
        mT5 & 24.24 \\
        \quad \textit{w.} cont-5langs & 24.22 \\
        \quad \textit{w.} Dr  & 23.75\\
        \midrule
        \multicolumn{2}{c}{\textit{Cross-Lingual}} \\
        \midrule
        \quad \textit{w.} DocNMT & 26.17\\
        \quad \textit{w.} DocTLM & 25.87 \\
        \systemnameone &  28.97 \\
        \systemnameone-Large & 30.52 \\
        \systemnametwo &  30.99 \\
        \systemnametwo-Large & \textbf{31.40} \\
        \bottomrule
    \end{tabular}
        \caption{Unseen language pair results on IWSLT 2015 Zh-En. Chinese is in the second-stage pretraining language set of \systemnametwo but not in those of \systemnameone. \systemnametwo-Large achieves SOTA.} 
    \label{tab:IWSLT2015}
\end{table}

\subsection{Document-Level Machine Translation} We evaluate \systemname on document translation. We split each document into chunks with a max length of 512 tokens. During inference, the decoded chunks are concatenated together to form the final document. We use prefix \textit{"Translate X to Y"} for translation, where X and Y are the source and target language names respectively.
\subsubsection{Seen Language Pair: WMT20 De-En}
WMT20 De-En is a document-level machine translation task. We use parallel training data from WMT20 without using additional monolingual data. From the results in Table ~\ref{tab:DR-WMT20}\footnote{For all the document translation experiments in this paper, the numbers are calculated using sacreBLEU \url{https://github.com/mjpost/sacrebleu} in document level.}, we see that \textit{Dr} provides large gains. \textit{DocNMT} outperforms \textit{DocTLM}. This is probably due to the fact that \textit{DocNMT} is more close to the document-level translation task. \textit{\systemnameone} once again outperforms Dr and other strong cross-lingual baselines. \textit{\systemnameone} is better than \textit{\systemnametwo} again because of capacity dilution as noted in \citet{aharoni2019massively}. As expected, \textit{\systemnameone-Large} outperforms \textit{\systemnameone} and to the best of our knowledge, achieves the SOTA. Note that previous systems use one or more of the following techniques: additional monolingual data, back-translation, ensembling or re-ranking tailored to a single translation pair.

\subsubsection{Unseen Language Pair: IWSLT 2015 Zh-En}
We use IWSLT 2015 Zh-En, another document-level machine translation task, to examine the multilingual transferability of \textit{\systemname} when the target transfer language (Chinese in this case) is of a very different script. Chinese is only in the first-stage pretraining of mT5 but not in our second-stage pretraining. We use parallel training data from IWSLT15 without using additional monolingual data. Following HAN \cite{werlen2018document}, we use 2010-2013 TED as the test set. The results are in Table \ref{tab:IWSLT2015}. \textit{\systemnameone} outperforms the strong cross-lingual and mono-lingual baselines, demonstrating impressive transfer capability . \textit{\systemnametwo} includes Chinese as one of the second-stage pretraining languages therefore obtains better numbers than \textit{\systemnameone}. Unsurprisingly, large models are better than their corresponding base models. To the best of our knowledge, \textit{\systemnametwo-Large} achieves the SOTA on this task. We qualitatively analyze the translations of different systems in Appendix \ref{sec:example}.

\begin{table}[ht]
    \small
    \centering

    \begin{tabular}{l|c|c|c|c|c}
    \toprule
    \textbf{Pretrained Model} &  \textbf{De-En} & \textbf{Ru-En}& \textbf{Pl-En} & \textbf{Ja-En} \\
        \midrule
        mT5\\
        \quad \textit{w.} DocNMT & 44.09 & 40.48& 3.13 & 0.92 \\
        \quad \textit{w.} DocTLM & 0.31 & 0.11 & 0.23 & 0.22 \\
        \midrule
        \systemnameone & 21.74 & 15.84 & 2.81 & 0.47 \\
        \systemnameone-Large & 35.63  & 29.50 & 14.15 & 1.16 \\ 
        \systemnametwo & 22.00 & 14.62 & 17.40 & 16.93 \\
        \systemnametwo-Large & 28.24 & 24.34 & 23.18 & 19.17\\
        \bottomrule
    \end{tabular}
        \caption{Document translation without finetuning on WMT20 De-En, Ru-En, Pl-En and Ja-En.} 
    \label{tab:zeroshot-WMT20}
\end{table}

\vspace{-0.4cm}
\subsubsection{Document Translation Without Finetuning}
We further show that \textit{\systemname} is able to perform document translation without finetuning, i.e., evaluate the model right after second-stage pretraining without any finetuning on task-specific data. We show the results in Table \ref{tab:zeroshot-WMT20}. While the mono-lingual pretrained models completely fail to produce meaningful translations, \textit{\systemnameone} is able to achieve over 20 BLEU points in De-En and 15 in Ru-En. Not surprisingly, \textit{\systemnameone-Large} further improves  to over 35 and 29 respectively. \textit{\systemnametwo} includes Pl-En and Ja-En in the second-stage pretraining and therefore obtains competitive results on these two language pairs with either base or large model. Although \textit{\systemnameone} is not pretrained on Pl-En, the large model gets over 14 BLEU on this task. One hypothesis is that Polish uses the Latin script and shares common subwords with German and Spanish, allowing our model to transfer knowledge across languages. On the other hand, the \textit{\systemnameone-Base} model fails to produce meaningful translations for Pl-En. This shows the importance of size when performing multilingual pretraining. The best model is \textit{DocNMT} which obtains over 40 BLUE points in both De-En and Ru-En, outperforming \textit{\systemnameone} and \textit{\systemnametwo}. This is reasonable because \textit{\systemname} shuffles documents in pretraining and this is misaligned with the document translation task inputs. The impressive performance of both \textit{DocNMT} and \textit{\systemname} shows that our MTmC4 corpus is of very high-quality and is likely better than the parallel data provided by the specific tasks in question. Further analysis of the quality of this data will be an interesting avenue for future work.

\section{Analysis}
\label{sec:analysis}
\subsection{Are Document-level Models Better Than Sentence-level Models?}
\begin{figure}[t!]
    \centering
    \includegraphics[width=\linewidth]{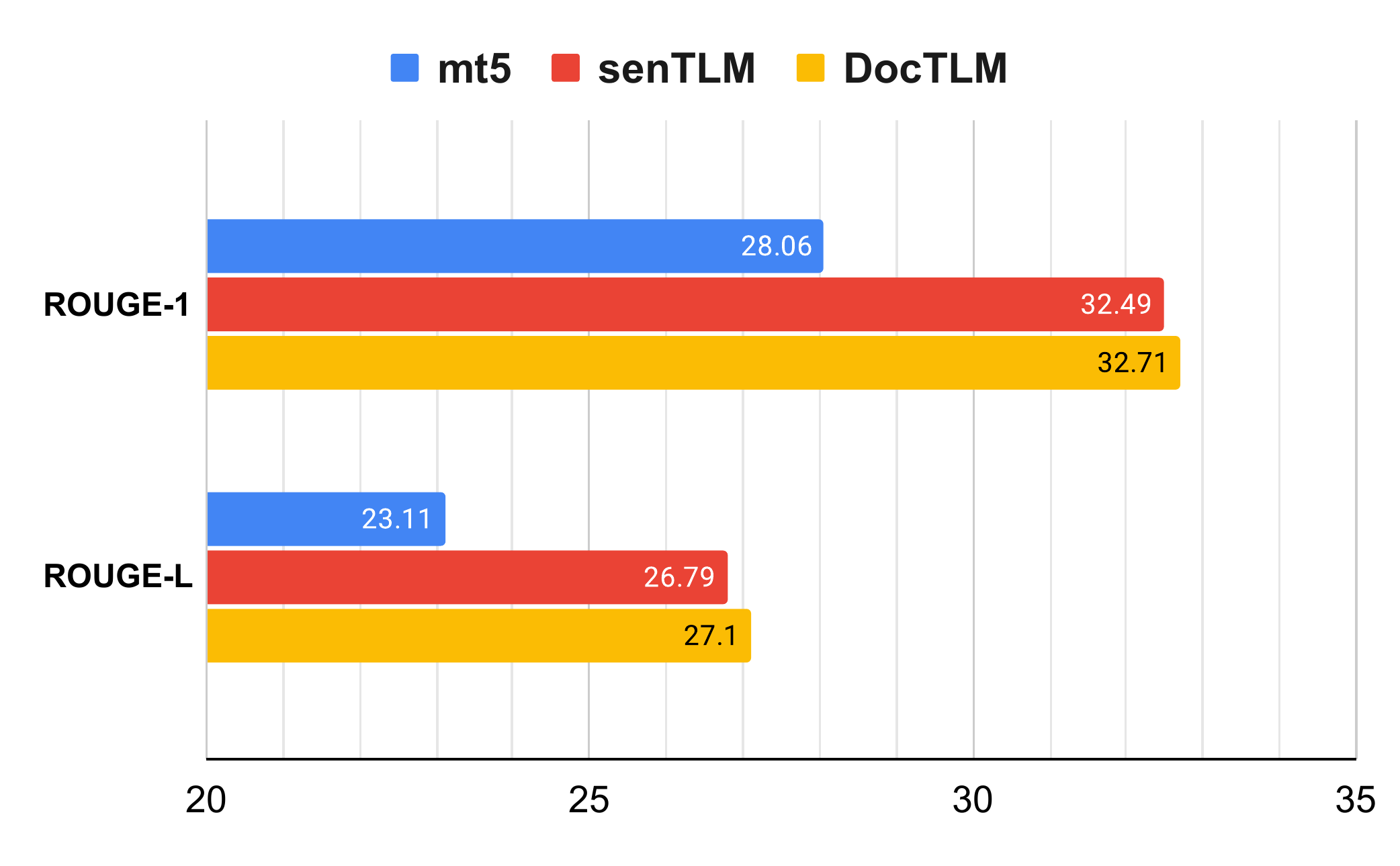}
    \caption{SenTLM and DocTLM finetuning results on Wikilingua. The numbers are average of four languages: \{Es, Tr, Ru, Vi\}.}
    \label{fig:DocSenWiki}
\end{figure}

To demonstrate the benefits of pretraining with longer context, we pretrain mT5 using translation language modeling (TLM) on five languages: \{De, Es, Tr, Vi, Ru\} with two different inputs. In \textit{DocTLM}, we concatenate the parallel documents into a single training sequence. As for \textit{SenTLM}, we break down the document into individual sentences and find the alignments in the parallel document pair. Then we concatenate the single aligned sentence pair as a training sequence. We finetune these second-stage pretrained models on Wikilingua and WMT20 De-En. The results are shown in Figure \ref{fig:DocSenWiki} and Table \ref{tab:docVSsen-WMT}. We see that document-level models offer small improvements on summarization and very significant improvements on document-level translation, showing that the longer context is indeed useful.

\begin{table}[h]
    \small
    \centering

    \begin{tabular}{l|@{\hskip4pt}c}
    \toprule
    \textbf{Pretrained-Model} &  \textbf{BLEU}  \\
        \midrule
        mT5 & 29.08\\
        \quad \textit{w.} SenTLM & 34.68\\
        \quad \textit{w.} \textbf{DocTLM} &\textbf{37.74} \\
        \bottomrule
    \end{tabular}
        \caption{SenTLM and DocTLM finetuning results on WMT20 De-En.} 
    \label{tab:docVSsen-WMT}
    \end{table}

\subsection{Effect of Data Quality in Second-stage Pretraining}
\begin{figure}[t!]
    \centering
    \includegraphics[width=\linewidth]{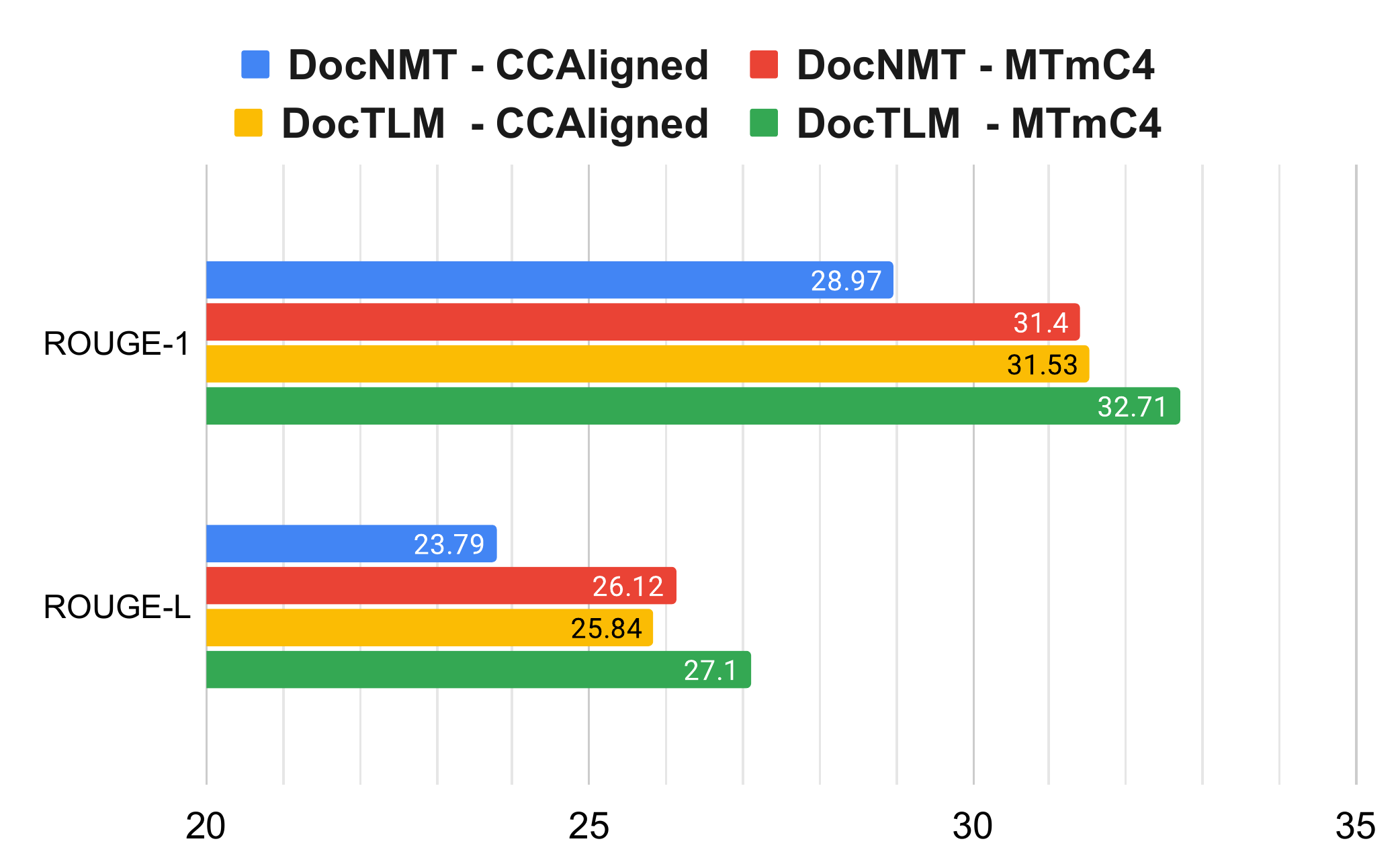}
    \caption{MTmC4 and CCAlgined finetuning results on Wikilingua. The numbers are average of four languages: \{Es, Tr, Ru, Vi\}.}
    \label{fig:data-wiki}
\end{figure}

\begin{figure}[t!]
    \centering
    \includegraphics[width=\linewidth]{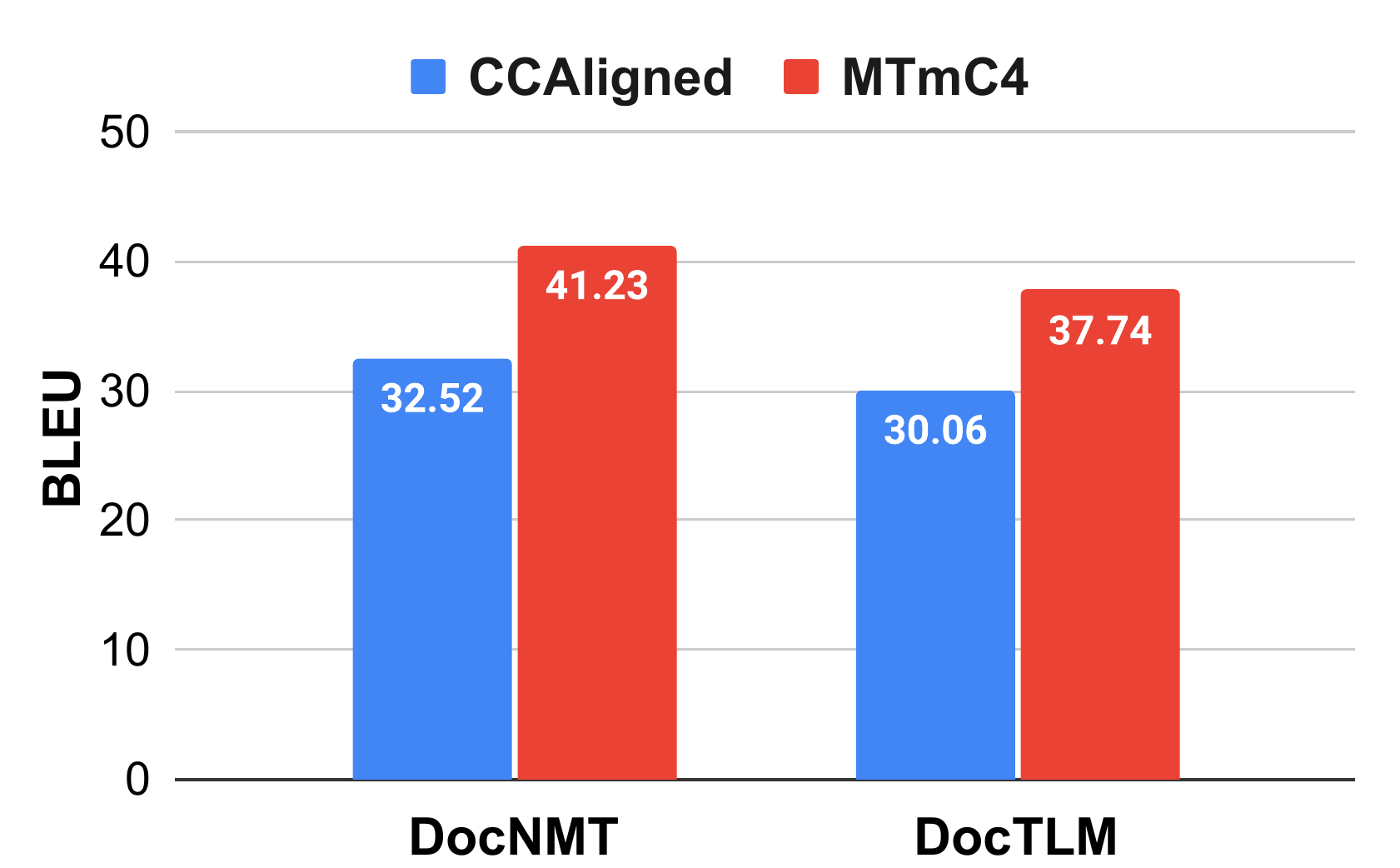}
    \caption{MTmC4 and CCAlgined finetuning results on WMT20 De-En.}
    \label{fig:data-WMT20}
\end{figure}

In our experiments, we observe big differences between different parallel corpora. We compare against the CCAligned corpus -- a large automatically mined corpus from Common Crawl which is found to be very noisy \cite{kreutzer2021quality}. In contrast, MTmC4 is produced by using high-quality translation systems. We pretrain mT5-Base on five languages: \{De, Es, Tr, Vi, Ru\} with these two corpora using \textit{DocNMT} and \textit{DocTLM}. We demonstrate the Wikilingua results in Figure \ref{fig:data-wiki} and WMT20 De-En results in Figure \ref{fig:data-WMT20}. Using our curated MTmC4 is consistently better regardless of pretraining objectives or tasks. 

\subsection{Does Combining Mono-Lingual and Cross-Lingual Pretraining Help?}
Here we try to see if combining both monolingual and cross-lingual objectives helps. We try two different continual pretraining strategies for combining Dr and \targetname. We use five languages: \{De, Ru, Tr, Vi, Es\}.
\textbf{(i)} Dr $\rightarrow$ \targetname: We first pretrain mT5 with Dr on mC4 for 0.5M steps and then pretrain with \targetname on MTmC4 for 0.5M steps. \textbf{(ii)} Dr + \targetname: We mix these two objectives with a 50-to-50\% ratio and pretrain for 0.5M steps. In Table \ref{tab:sequential-WMT20}, we show that \textbf{(i)} slightly improves over only \targetname in both tasks and \textbf{(ii)} slightly improves on WMT20 De-En but seems to hurt performance on ISWLT15 Zh-En.

\begin{table}[h]
    \small
    \centering

    \begin{tabular}{l|@{\hskip3pt}c@{\hskip3pt}c@{\hskip3pt}c}
    \toprule
    \textbf{Pretrained-Model} &  \textbf{WMT20 De-En} &  \textbf{IWSLT15 Zh-En} \\
        \midrule
        mT5  \\
        \quad \textit{w.} Dr & 36.63 & 23.75 \\
        \quad \textit{w.} \targetname   & 42.05 & 28.00 \\
        \quad \textit{w.} \textbf{Dr} $\rightarrow$ \textbf{\targetname} & \textbf{42.75} & \textbf{28.18} \\
        \quad \textit{w.} Dr + \targetname & 42.37 & 27.35 \\
        \bottomrule
    \end{tabular}
        \caption{Methods of combining mono-lingual and cross-lingual and their finetuning results on WMT20 De-En and IWSLT15 Zh-En.} 
    \label{tab:sequential-WMT20}
\end{table}

\subsection{How Many Pretraining Steps is Required for \targetname?}
\begin{figure}[t!]
    \centering
    \includegraphics[width=\linewidth]{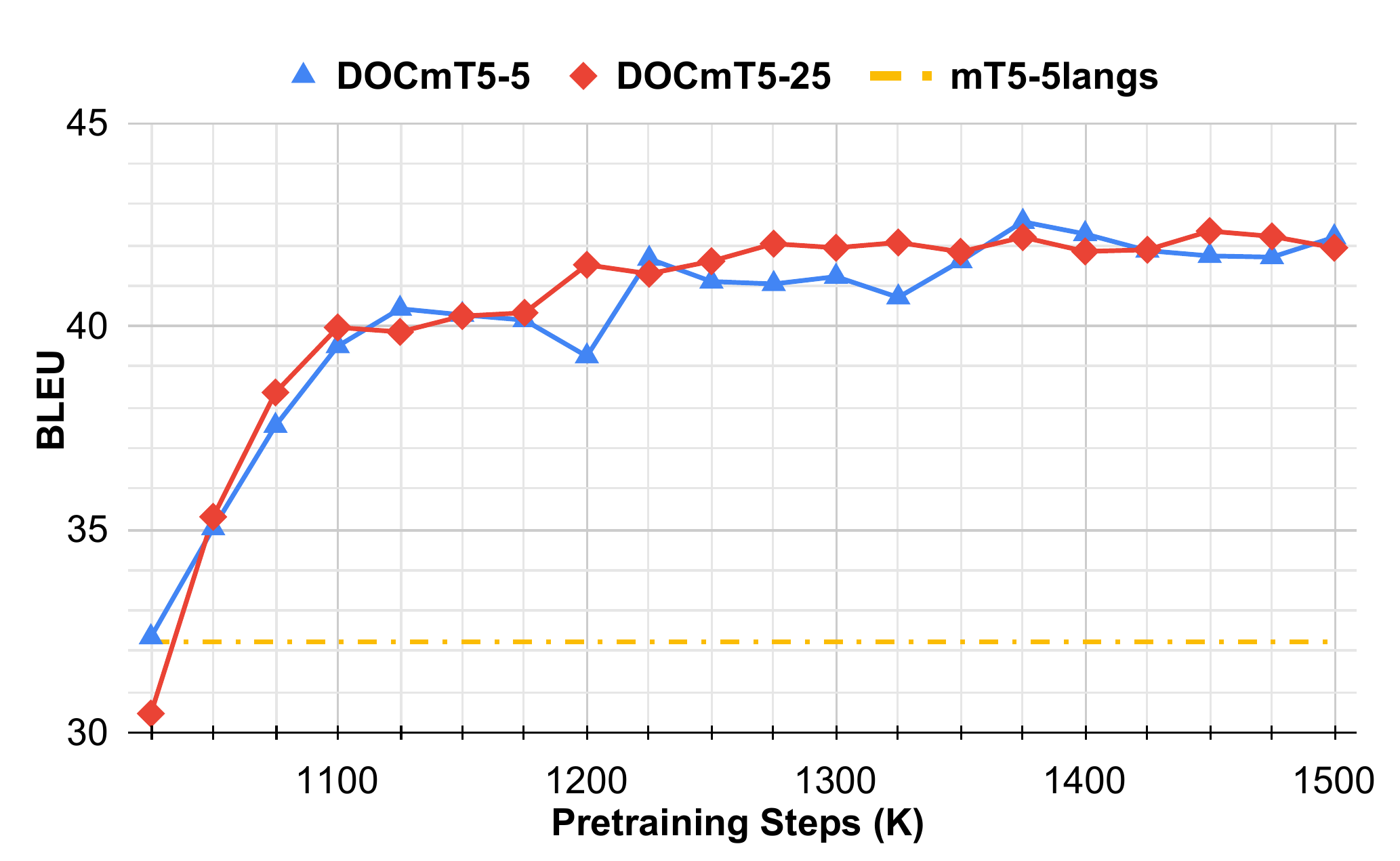}
    \caption{finetuning results of WMT20 De-En along with pretraining steps. We use \systemnameone-base.}
    \label{fig:steps}
\end{figure}

To answer this question, we take different pretraining checkpoints of \textit{\systemnameone} and \textit{\systemnametwo} and finetune with WMT20 De-En. The results are shown in Figure \ref{fig:steps}. After 50k steps of pretraining with \textit{DrMT}, both systems outperform the \textit{cont-5langs}. After 300k steps, both systems roughly converge and perform similarly.  

\section{Conclusion}
In this paper, we present \systemname, a novel document-level multilingual pre-trained model. Our proposed objective, \targetname, is simple and effective and leads to large gains over strong baselines (e.g. mBART and MARGE) on cross-lingual summarization and document-level translation. \systemname achieved SOTA on two competitive document-level translation tasks: WMT20 De-En and IWSLT15 Zh-En. We further analyze various factors that contribute to successful document-level pre-training. We plan to release the pre-trained model to facilitate future work on document-level language understanding.

\section*{Acknowledgements}
We would like to thank Alexis Conneau, Jon Clark and Mihir Sanjay Kale for the helpful discussions. We also thank Sebastian Ruder, Noah Constant and Ankur Bapna for providing feedback on the manuscript.

\bibliography{anthology,custom}
\bibliographystyle{acl_natbib}

\clearpage
\appendix

\appendix
\appendixpage
\begin{figure*}[b]
    \centering
    \includegraphics[width=\linewidth]{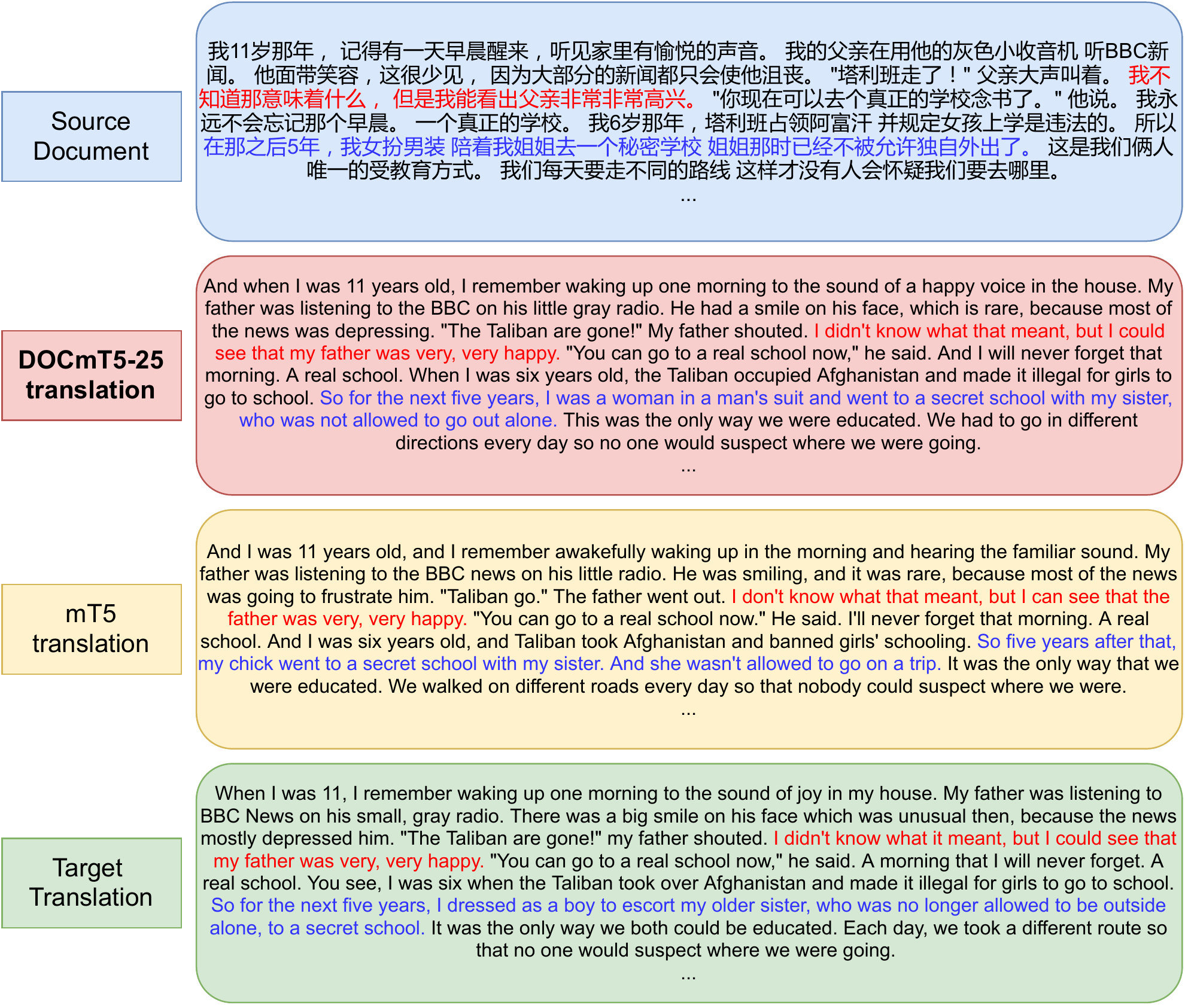}
    \caption{A comparison example of Zh-En document translation. \systemname is able to produce consistent time tenses while mT5 baseline fails. \systemname also produces longer and conherent sentences. Best viewed in color.}
    \label{fig:example}
\end{figure*}

\section{Analysis of Document Translation}
\label{sec:example}
We take a deeper look at the translations produced by various systems to understand what makes \textit{\systemname} better. We demonstrate an example in Table \ref{fig:example}. We take the best system (\textit{\systemnametwo-Large}) and the \textit{cont-5langs} baseline. We observe that \textit{\systemname} uses time tenses better than the baseline, producing more coherent sentences (red-colored texts). Additionally, \textit{\systemname} handles a compositional sentence more elegantly, instead of just using "and" (blue-colored texts). Finally, we observe that \textit{cont-5langs} often makes minor translation mistakes while our \textit{\systemname} makes much fewer of them. 

\end{document}